\documentclass[10pt,journal,compsoc]{IEEEtran}
\usepackage[nocompress]{cite}
\usepackage[pdftex]{graphicx}
\usepackage{amsmath}
\usepackage{algorithmic}
\usepackage{array}
\usepackage[caption=false,font=footnotesize,labelfont=sf,textfont=sf]{subfig}
\usepackage{fixltx2e}
\usepackage{stfloats}
\ifCLASSOPTIONcaptionsoff
  \usepackage[nomarkers]{endfloat}
 \let\MYoriglatexcaption\caption
 \renewcommand{\caption}[2][\relax]{\MYoriglatexcaption[#2]{#2}}
\fi
\usepackage{url}

\usepackage{booktabs}
\usepackage{amssymb}
\usepackage{makecell}
\usepackage{multirow}
\usepackage{bm}
\usepackage{comment}
\usepackage{pifont}
\newcommand{\cmark}{\ding{51}}%
\newcommand{\xmark}{\ding{55}}%

\newcommand{\eg}{e.g.}

\newcommand{\etal}{et al.}

\begin{document}
\title{Recent Advances in Large Margin Learning}

\author{Yiwen Guo and Changshui Zhang,~\IEEEmembership{Fellow,~IEEE}
 \IEEEcompsocitemizethanks{
  \IEEEcompsocthanksitem Y. Guo is with ByteDance AI Lab, Beijing 100000, China. E-mail: guoyiwen.ai@bytedance.com.
  \IEEEcompsocthanksitem C. Zhang is with the Institute for Artificial Intelligence, Tsinghua University (THUAI), Beijing National Research Center for Information Science and Technology (BNRist), and the Department of Automation, Tsinghua University, Beijing 100084, China. E-mail: zcs@mail.tsinghua.edu.cn}
 \thanks{Manuscript received Oct. 20, 2019, revised Mar. 20, 2021.}}



\IEEEtitleabstractindextext{
 \begin{abstract}
This paper serves as a survey of recent advances in large margin training and its theoretical foundations, mostly for (nonlinear) deep neural networks (DNNs) that are probably the most prominent machine learning models for large-scale data in the community over the past decade. We generalize the formulation of classification margins from classical research to latest DNNs, summarize theoretical connections between the margin, network generalization, and robustness, and introduce recent efforts in enlarging the margins for DNNs comprehensively. Since the viewpoint of different methods is discrepant, we categorize them into groups for ease of comparison and discussion in the paper. Hopefully, our discussions and overview inspire new research work in the community that aim to improve the performance of DNNs, and we also point to directions where the large margin principle can be verified to provide theoretical evidence why certain regularizations for DNNs function well in practice. We managed to shorten the paper such that the crucial spirit of large margin learning and related methods are better emphasized. 
 \end{abstract}

 }

\maketitle
\IEEEdisplaynontitleabstractindextext
\IEEEpeerreviewmaketitle

\IEEEraisesectionheading{\section{Introduction}\label{sec:intro}}
\IEEEPARstart{T}{he} concept of large margin learning arises along with the development of support vector machine (SVM)~\cite{Cortes1995, Vapnik1999}, which aims to fix the empirical risk and minimize the confidence interval, in contrast to many models that target mostly at minimizing the empirical risk~\cite{Vapnik1999, Vapnik2013}. 
Benefit from solid theoretical basis from statistical learning, large margin classifiers show promise in both generalization ability and robustness. Since the late 1990s, they have been intensively studied and widely adopted~\cite{Crammer2001, Tsochantaridis2005, Zhu2004, Cotter2013, Smola2000}. 

Yet, every learning machine has its day. Recent years have witnessed a revive of neural networks, partially owing to the advances of computational units that are capable of processing large-scale datasets. By learning representations from data, they, especially deep neural networks (DNNs), have advanced the state-of-the-arts of many tasks for machine intelligence~\cite{He2016, Graves2013}.
One might be curious about the relationship between DNNs and conventional large margin classifiers (e.g., SVM), and in view of this, we would like to answer three questions in this survey: \textbf{1)} Is the large margin principle essential or at least beneficial to the classification performance of DNNs? \textbf{2)} if yes, is it (implicitly) supported with a normal training mechanism used in practice? \textbf{3)} If not implicitly supported, how to gain large margins for classification models that are nonlinear and structural complex like DNNs? This paper presents an overview of existing work on these points. To the best of our knowledge, there is no such survey in the literature, and our work will bridge the information gap and inspire new research hopefully.

\subsection{A Formal Definition of Classification Margin}
To get started, let us first introduce a formal definition of the \textbf{classification margin}, which applies to a variety of different classifiers, including both linear and nonlinear ones. 

It can be slightly different to formulate the margin of binary and multi-class classification. We first consider the binary scenario, in which a label $y$ from $\{+1, -1\}$ is assigned by a classifier to its input $\mathbf x \in \mathbb R^n$.
Given $f:\mathbb R^n\rightarrow \mathbb R$ which maps the $n$-dimensional input $\mathbf x$ into a one-dimensional decision space, an \emph{instance-specific margin} is defined as 
\begin{equation}
m_{\mathbf{x}, f}:=\min \|\mathbf z\|_p \ \ \mathrm{s.t.}\ f(\mathbf x)f(\mathbf x+\mathbf z)<0
\end{equation}
for any $\mathbf x\in \mathbb R^n$. $p\geq1$ indicates the concerned $l_p$-norm, and, in the Euclidean space, we have $p=2$. For a classifier whose prediction is given by a linear function of its input $\mathbf x$, we can rewrite the function output $f(\mathbf x)$ as $f(\mathbf x)=\mathbf w^T \mathbf x+b$, and the instance-specific margin has a closed-form solution in this case: $m_{\mathbf{x}, f}=|f(\mathbf x)|/\|\mathbf w\|$. 
It can be of great interest to study a margin $m_f:=\min m_{\mathbf{x}, f}$ over the whole training set.
SVM in the linear case enlarges such a classification margin by minimizing $\|\mathbf w\|$ and constraining the value of $f(\mathbf x)$ (to be more specific, by letting $yf(\mathbf x)>1$ for all $\mathbf x \in \mathbb R^n$). 
Taking advantage of some kernel trick~\cite{Shawe2004} and utilizing nonlinear functions in reproducing kernel Hilbert spaces, it is natural to extend the expressions and analyses of margins to SVMs with higher expressivity~\cite{Vapnik2013}. 

With the expression of classification margin provided for SVMs, we obtain certification on the superiority of their generalization ability and robustness~\cite{Bartlett1999, Scholkopf2000, Xu2009}.
For DNNs, it is very challenging, if not impossible, to derive analytic expressions for their classification margins, on account of the hierarchical model structure and complex nonlinearity of $f$. 
Therefore, the study of classification margin, generalization ability, and robustness of DNNs also attract great attention recently. 
In the following sections, we will try to answer the three questions and given an overview of recent advances in margin related studies for DNNs. 
Before providing more details, it is worth stressing that the ``classification margin'' defined in the beginning of this section and concerned in linear SVMs is in the \emph{input space} of learning models. We will also mention some margins in the output or representation spaces of DNNs in this paper.

\section{The Relationships between Margin, Generalization, and Robustness}

In this section, we attempt to answer the first two questions raised in Section~\ref{sec:intro}. First, it is the question about benefits of large margin learning to DNN-based classifications.
Related studies from theoretical perspectives have been performed for decades on the basis of some shallow models like SVMs. In this section, we focus on research work based on DNNs. 

\subsection{Large Margin Is Beneficial to DNNs as Well}
Before delving deep into these studies, we first introduce the concept of \textbf{robustness}, (a.k.a., algorithmic robustness~\cite{Xu2012} and robustness in patterns~\cite{Smola2000}), which is an essential property of classifiers and closely related to the classification margin.
As is known from the definition, for any reasonable input (e.g., any natural image that can be fed into a scene classification system), the classifier will hold its prediction with any pixel-wise perturbation smaller than the margin (i.e., $\|\mathbf z\|_p<m_f$).
We consider $f$ as a robust model if $m_f$ is sufficiently large so that the perturbation lead to samples perceptually from the other classes.
Over the last few years, the robustness of DNN models has draw more attention along with developments of adversarial attacks~\cite{Szegedy2014, Goodfellow2015, Moosavi2016, Kurakin2017, Xie2017}. It has been demonstrated that one can easily manipulate the prediction of a state-of-the-art DNN model by adding subtle perturbations to its input. By definition, given the function $f$ and an input $\mathbf x$, any perturbation to $\mathbf x$ within the hyper-sphere $\{\mathbf z|\|\mathbf z\|< m_{\mathbf{x}, f}\}$ would not alter the model prediction.
That said, the concept of {adversarial robustness} that describes the ability of a model to resist adversarial attacks, is intrinsically related to the classification margin. 

\begin{figure}[h]
\begin{center}
\includegraphics[width=0.42\textwidth,clip]{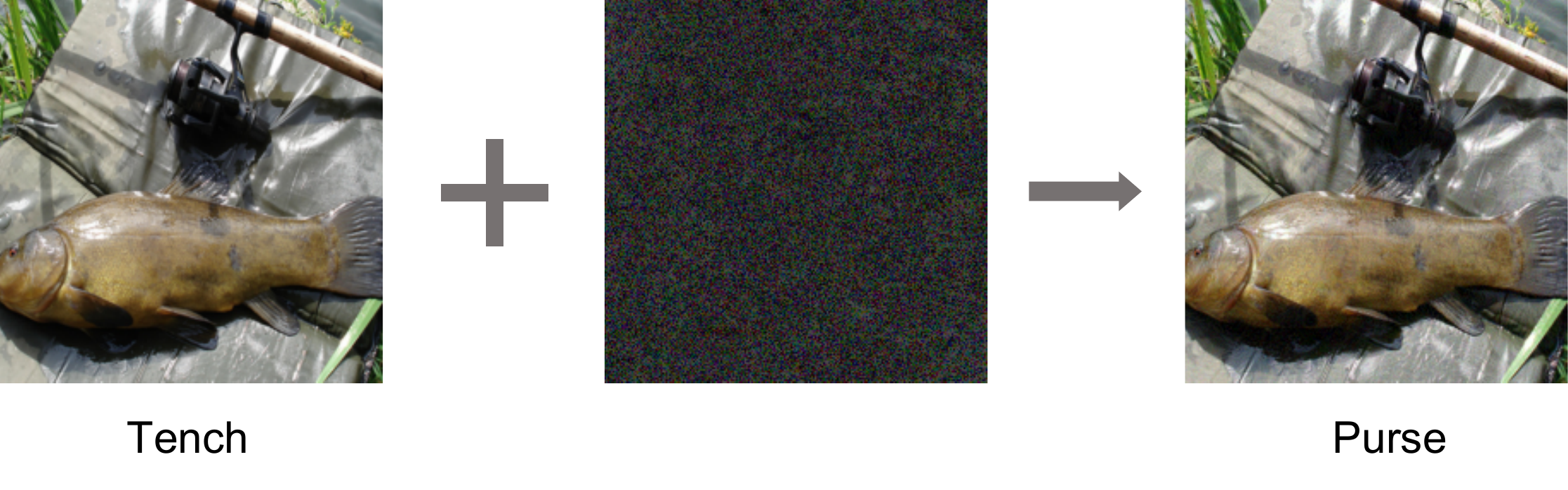} \vskip -0.1in
\caption{An example of the adversarial example which is misclassified as purse by a ResNet-50 trained on ImageNet. We enlarge the perturbation by $10\times$ for better illustration in the middle picture.}
\label{fig:relationship}
\end{center}
\vskip -0.05in
\end{figure}

What relates to robustness and being important as well is the \textbf{generalization ability}.
Suppose that a DNN model is to be learned to minimize the risk $\tilde l(\Theta):=\mathbb E_{\mathbf x, y}(l(f(\mathbf x; \Theta),y))$, in which $\Theta$ is a set that exhaustively collects all learnable parameters in the network. Since the joint distribution of $\mathbf x$ and $y$ is generally unknown in practice, the objective of expected risk minimization cannot be pursued directly and we opt to minimizing an empirical risk instead, using a set of $N$ training samples $\{(\mathbf x_i, y_i)\}$, i.e., $l_{emp}(\Theta):=\mathbb \sum_i l(f(\mathbf x_i; \Theta),y_i)/N$. In spite of being a rule of thumb, there is always a gap between minimizing $\tilde l(\Theta)$ and $l_{emp}(\Theta)$, and we can use the generalization error $\mathrm{GE}(g)=|\tilde l(\Theta)-l_{emp}(\Theta)|$ to characterize such a gap formally~\cite{Xu2009, Kawaguchi2017}. 

It has been discovered that the margin and some other robustness measure of a classifier bound the generalization error of the classifier in theory~\cite{Xu2009, cao2019learning}. 
Naturally, we prefer machine learning models with lower generalization error and it has been demonstrated that, for a $(K, \epsilon (\mathcal S_n))$-robust algorithm with $l(f(\mathbf x, :), y)\leq M$ for all reasonable $(\mathbf x, y)$~\footnote{Formal definition of $(K, \epsilon (\mathcal S_n))$ can be found in~\cite{Xu2009, Sokolic2017} and it characterizes how the training data is exploited by the algorithm. $K$ is the number of sample partitions and $\epsilon (\mathcal S_n)$ bounds the discrepancy between training and possible test losses in each partition.}, we have, with probability at least $1-\delta$, it holds that
 \begin{equation}
  \mathrm{GE}(g)\leq \epsilon(\mathcal S_n)+M\sqrt{\frac{2K\log(2)+2\log(1/\delta)}{N}}.
 \end{equation}
The result establishes connections between the generalization ability and robustness of classification models.
Inspired the result, Soboli\'c et al.~\cite{Sokolic2017} proposed to bound the margin of a DNN such that both its robustness and generalization ability are guaranteed, and it was achieved by constraining the Jacobian matrix. 
For a classification model that was fed with an $n$-dimensional input each time and outputted $k$ neurons before softmax, the $n\times k$ Jacobian matrix should also be instance-specific and it was obtained by calculating the gradient of the function with respect to its input. 
Superior test-set accuracies were obtained using the Jacobian regularization.
They also generalized the theoretical and empirical analyses to stable invariant classifiers (e.g., convolutional neural networks, CNNs)~\cite{Sokolic2017b, huang2015}, and methods that could enhance robustness to data variations are suggested. 
With all the facts, we know that, for Question \textbf{1}, the large margin principle is beneficial to DNNs, in improving the generalization ability and robustness.

\subsection{Large Margin Cannot be Trivially Obtained}
Let us now turn to answering Question \textbf{2}. Apparently, large margins cannot be naturally obtained with a normal training mechanism (i.e., using stochastic gradient descent to minimize just a cross-entropy loss) for DNN models in practice, otherwise their adversarial vulnerability would not be considered severe. 
Note that although it has been proved that {linear networks} trained on \emph{separable data} using stochastic gradient descent converge to maximum margin solutions as $t\rightarrow \infty$~\cite{Soudry2018, Gunasekar2018, Ji2019, Nacson2019}, it was also demonstrated that the convergence rate was extremely slow (e.g., $O(1/\log(t))$~\cite{Soudry2018} using the cross-entropy loss), making it hardly achievable in practice.
Similar results can also be derived for \emph{non-separable data}~\cite{ji2019implicit}.
In addition to the results that were derived on the basis of the implicit bias of stochastic gradient descent, it has been shown that over parameterization also leads to improved margins~\cite{Wei2019}. 
Under an infinite network width regime, stochastic gradient descent of a two-layer network model leads to an inference function in a reproducing kernel Hilbert space of the neural tangent kernel~\cite{daniely2017sgd, jacot2018neural, arora2019exact}, and it can be proved that a proper explicit regularization can indeed improve margins.

\section{Achieve Large Margin for DNNs}\label{sec:lmdnn}

Now that we have answered the first two concerned questions in the previous section, we will attempt to answer the third question in this sections. Figure~\ref{fig:relationship} is a summarization of what follows. Section~\ref{sec:lip_outmargin} to~\ref{sec:certif} attempt to group training mechanisms that affect the margin with theoretical guarantees into several categories for better clarity.

\begin{figure}[h]
\begin{center}
\includegraphics[width=0.42\textwidth,clip]{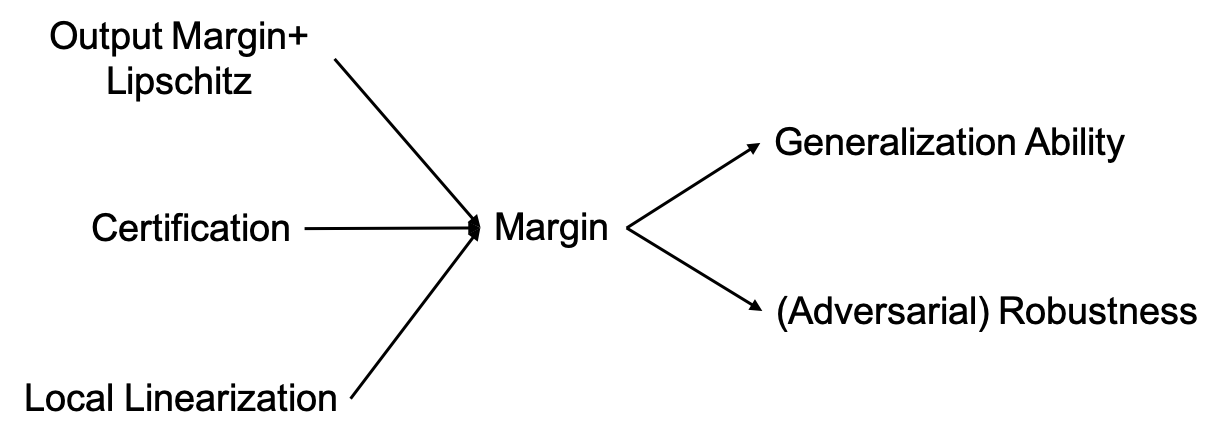} \vskip -0.1in
\caption{The concerned classification margin is closely related to adversarial robustness and generalization ability, and we will introduce methods that aim to enlarge the margins for DNNs, by local linearization~\cite{Yan2018, Yan2019, Elsayed2018}, certification~\cite{Tjeng2017, Wong2018, Cohen2019}, or relying on a margin in the decision space~\cite{Liu2016, Wang2018cosineface, Deng2019} in combination with some (possibly implicit) Lipschitz constraints.}
\label{fig:relationship}
\end{center}
\vskip -0.05in
\end{figure}

\subsection{Regularization on Lipschitz Constant and Output Margins}\label{sec:lip_outmargin}
We know that conventional regularization strategies like weight decay~\cite{Krogh1992} and dropout~\cite{Hinton2012} do not help in enlarging the margin for nonlinear DNNs, which seems different from SVMs.
It indicates that the hierarchical structure of DNNs is worthy further exploring, considering that the optimization of an SVM can be roughly considered as training the final layer of a DNN with weight decay. 

Unlike for SVMs, derivation and calculation of a classification margin seem infeasible for nonlinear DNNs, and hence some approximations are pursued as surrogates. For instance, as mentioned, Soboli\'c et al.~\cite{Sokolic2017} took advantage of the fact that nonlinear DNNs were Lipschitz continuous and used the inequality
\begin{equation}\label{eq:lip}
\|f(\mathbf x)-f(\mathbf x')\|_q \leq L_{p,q}\|\mathbf x-\mathbf x'\|_p
\end{equation}
to constrain the margin, which was basically an $l_p$ distance between two points in the input space.
We know from Eq.~(\ref{eq:lip}) that the $l_q$ distance between any two points in the prediction space is an essential factor of the classification margin. Fix $\|f(\mathbf x)-f(\mathbf x')\|_q\leq \epsilon, \forall\mathbf x'$, the classification margin can be guaranteed by minimizing a~\textbf{Lipschitz constant} (i.e., $L_{p,q}$). 
It was also theoretically proven that the Lipschitz constant has direct relationships with both robustness~\cite{Hein2017, Weng2018} and generalization~\cite{Bartlett2017, Neyshabur2017, Wei2019data} for a variety of nonlinear DNNs. 
Unsurprisingly, the Lipschitz constant and some ``margins'' in the model decision space, which laid the foundation of a spectrum of methods as will be introduced in the following paragraphs, jointly guarantee the (adversarial) robustness of DNNs.
The norm of the Jacobian matrix, is then suggested to be penalized during training for improved robustness~\cite{Sokolic2017, Jakubovitz2018}, since it is actually a local Lipschitz constant fulfilling the inequality in Eq.~\eqref{eq:lip} for any specific $\mathbf x$ and it measures the sensitivity of a learning machine by definition~\cite{Novak2018}. 
In addition to the norm of the Jacobian matrix, the Euclidean norm of input gradient~\cite{Ross2018}, the curvature of Hessian matrix~\cite{Moosavi2019}, the spectral norm of weight matrices~\cite{Yoshida2017}, and a cross-Lipschitz functional~\cite{Hein2017} can all be used to enhance the robustness or generalization ability of DNNs, and larger margins are simultaneously anticipated. 
In fact, these network properties are all closely related to the Lipschitz constant and have some chained inequality~\cite{Jakubovitz2018,guo2020connections}. 
In binary classification, some of the regularizations share the same essential ingredients (i.e., maximizing prediction confidence and minimizing local Lipschitz constants) in theory and show similar empirical results~\cite{guo2020connections}. 

As mentioned, there is also a line of work focusing on ``margins'' in the decision or a representation space (i.e., \textbf{output margin}) of DNNs.
For instance, Tang~\cite{Tang2013} proposed to replace the softmax cross-entropy loss with an SVM-derived one, leading to margin gain in the representation space characterized by the penultimate layer and a winning solution to the ICML'13 representation learning challenge~\cite{Goodfellow2013challenges}. 
Sun \etal~\cite{Sun2016} introduced margin-based penalties to the objective of training DNNs, motivated by theoretical analyses from the perspective of the margin bound.
The triplet loss~\cite{Weinberger2009} that imposes a margin of representation distance between each positive sample pair and each negative sample pair was also considered, e.g., in FaceNet~\cite{Schroff2015}.
Since then, such output margins have been actively discussed in the face recognition community. 
Beside what was applied in FaceNet, angular softmax (A-Softmax)~\footnote{See also large margin softmax (L-Softmax) in~\cite{Liu2016}, which is very similar to A-Softmax.} and additive margin softmax (AM-Softmax) were developed and used in SphereFace~\cite{Liu2017} and CosineFace~\cite{Wang2018cosineface}, respectively, to enhance the cross-entropy loss with novel softmax formulations.
There are also ensemble soft-margin softmax (M-Softmax)~\cite{Wang2018}, virtual softmax (V-Softmax)~\cite{Chen2018}, large margin cosine loss (LMCL)~\cite{Wang2018cosface}, and additive angular margin loss (AAML)~\cite{Deng2019}, just to name a few.
They have achieved remarkable success in the task of face identification and verification, and some of them also showed promising accuracy for classifying natural scene images.
The difference between these methods lie in the way of decomposing the cross-entropy loss. Table~\ref{tab:face} summarizes and compares them. 
It is also worth mentioning that the cross-entropy loss itself can also be interpreted as a margin-based loss, with input-specific margins in the output space, and it was shown that enlarging such input-specific margins could be used as a regularization and led to improved test-set accuracy~\cite{kobayashi2019large}.

Similar ideas for enlarging output margins have also been considered in the task of speech recognition~\cite{Zhang2015speech, Zhang2016recurrent, Wang2019large}, in which computational modules like feedback connections~\cite{Graves2013} and self-attentions~\cite{Bahdanau2014} often serve in the backbone models. 
It is expected to obtain deep feature representations that lead to the largest SVM margins.
A two-stage pipeline was initially developed for training such models~\cite{Zhang2015speech, Zhang2016recurrent}, in which learnable parameters in the final and prior layers were updated separately. 
For more recent methods, automatic differentiation~\cite{Paszke2017} was adopted, such that the feature representations and large margin SVM classifiers can be optimized jointly. Such a large margin principle was also utilized in few-shot learning~\cite{Wang2018few}, PU learning~\cite{Elkan2008, Gong2018pu}, and anomaly detection~\cite{Liu2019anormaly}.

\begin{table}[!t]
 \caption{Large margin classification for face recognition. The methods differ from each other by introducing margins in angular, logit, or cosine spaces and by incorporating scaling factors (Mul) or additive terms (Add). Given $W$ as a learnable matrix before softmax and $h$ as the feature representation of a DNN, if $\|W\|_2=1$ is fixed, then we can rewrite the linear transformation $W^Th$ as $\cos(\theta)\|h\|_2$ and incorporate a scaling factor $\alpha$ and an additive term $\beta$ into it to encourage the angular margin as $\cos(\alpha\theta+\beta)\|h\|_2$. Note that although M-Softmax targets to image classification, it is closely related to the other methods and thus we list it here as well.}\label{tab:face}
 \centering
 \begin{tabular}{lcccc}
  \toprule
           Method  & Test data    & Margin   & Add & Mul\\
  \midrule
  L-Softmax~\cite{Liu2016}         & image, face & Angular & \xmark &  \cmark      \\
  A-Softmax~\cite{Liu2017}          & face             & Angular & \xmark  & \cmark      \\
  M-Softmax~\cite{Wang2018}          & image         & Logit  & \cmark & \xmark      \\
  V-Softmax~\cite{Chen2018}          & image, face &  Logit & \xmark &\xmark     \\
  AM-Softmax~\cite{Wang2018cosineface}      & face             & Cosine & \cmark & \xmark     \\
  LMCL~\cite{Wang2018cosface}		     & face	      & Cosine & \cmark & \xmark     \\
  AAML~\cite{Deng2019}                & face	      	     & Angular & \cmark & \xmark     \\ \bottomrule
 \end{tabular}
\end{table}

While these methods considered classification margins of all training instances, it seems more efficient and reasonable to mainly focus on support vectors, just like in SVMs. In this spirit, Wang et al.~\cite{Wang2018support} combined the margin-based softmax with hard example mining~\cite{Shrivastava2016, Lin2017focal}, such that training focused more on harder samples which were considered more informative. Since only the ``support vectors'' were used for calculating gradients and updating parameters, the training process became more efficient. Being viewed as a functional abstraction of the training dataset, the set of ``support vectors'' has also been utilized to address catastrophic forgetting~\cite{Kemker2018} in incremental learning DNNs~\cite{Li2018supportnet}.

\subsection{Local Linearization}\label{sec:local_linear}
The methods introduced in Section~\ref{sec:lip_outmargin} enlarged ``margins'' mostly in the decision space of DNN models. 
Efforts might also be devoted in other representation spaces~\cite{Zhong2019}, however, 
as discussed~\cite{An2015}, owing to the distance distortions between input and representation spaces, the classification margins in the input space of DNNs were not necessarily maximized by methods in this category~\footnote{Although it has been theoretically shown that enlarging the output margin is beneficial to the generalization ability of DNNs, along with constrained classifier norms~\cite{Bartlett2017,neyshabur2017pac} or constrained complexity of each layer~\cite{wei2019improved} }.
An et al.~\cite{An2015} hence further enforced the transformations of middle layers of a DNN to be contraction mappings, in order to achieve large classification margins.
One step further, Bansal et al.~\cite{Bansal2018} advocated a similar learning objective to that of SVMs, in which the method of Lagrange multipliers is utilized.
Both of them can be seen as using layer-wise approximations to restricting some whole network property (e.g., the Lipschitz constant).
Recently, independent work from Yan et al.~\cite{Yan2018} and Elsayed et al.~\cite{Elsayed2018} proposed to enlarge classification margins via \textbf{local linearization} and reasonable approximations. 
In fact, by simply rewriting the constraint $f(\mathbf x)f(\mathbf x+\mathbf z)<0$ using Taylor's approximation of $f$ with respect to $\mathbf x$, one can obtain a closed-form solution to the worst-case perturbations even for nonlinear DNNs.
Elsayed et al. proposed to maximize 
\begin{equation}
\frac{|f(\mathbf x)|}{\|\nabla_{\mathbf x} f(\mathbf x)\|_q},
\end{equation}
in which $\|\cdot\|_q$ was the dual norm of $\|\cdot\|_p$, with $1/p+1/q=1$. 
In essence, the method can be considered as a simplified and efficient version of Yan et al.'s~\cite{Yan2018}, while Yan et al.'s method took one step further and followed a similar mechanism to DeepFool~\cite{Moosavi2016}. Specifically, an iterative local linearization was utilized by DeepFool and Yan et al.'s method to pursue approximations to evaluating the classification margins~\cite{Yan2018}, which can be more accurate yet more computational demanding.
Shortly after, Ding et al.~\cite{Ding2019} also discussed large margin training for DNNs, and they proposed to incorporate 
\begin{equation}
\sum_{i\in\mathcal S^+}\max \{0, m_{\max}-\hat{m}_{\mathbf x_i, f}\}
\end{equation}
into the learning objective of DNNs, in which $\hat{m}_{\mathbf x_i, f}$ was an estimation of the instance-specific margin $m_{\mathbf x_i, f}$ for $\mathbf x_i$, $\mathcal S^+$ indicated the set of correctly classified training samples using the network model, and $m_{\max}>0$ was a hyper-parameter. To well approximate $m_{\mathbf x, f}$, they took advantage of the PGD attack~\cite{Madry2018} and adopted its variant as a proxy of the ``shortest successful perturbation''. 

\begin{figure}[t]
 \centering
 \newcommand{\subfigwidth}{0.32\linewidth}
 \subfloat[min]{\includegraphics[width=\subfigwidth]{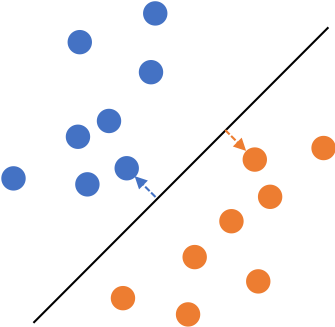}}\hskip 0.4in
 \subfloat[average]{\includegraphics[width=\subfigwidth]{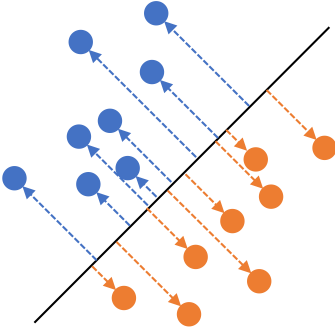}}\\
 \subfloat[aggregation]{\includegraphics[width=\subfigwidth]{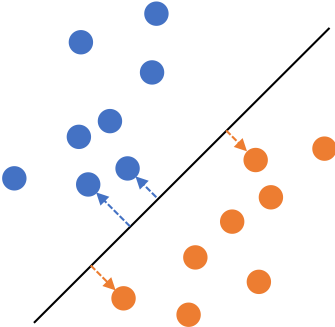}}\hskip 0.4in
 \subfloat[shrinkage]{\includegraphics[width=\subfigwidth]{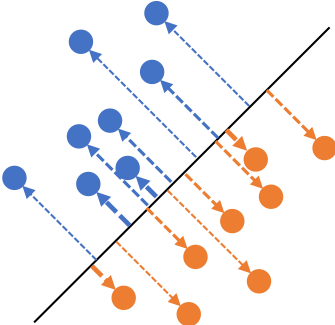}}
\caption{Different choices of of the key components result in different penalty on instance-specific margins. More specifically, (a) $\min$ considers only the worst case in the training set, (b) $\sum$ combined with an identity function treat all training samples equally, (c) and (d) incorporates the aggregation function and shrinkage function, respectively.}
\label{fig:regularizations}
\vskip -0.05in
\end{figure}

\subsubsection{Key Components and More Discussions}\label{sec:key}
In existing work that explicitly incorporates margin-based regularizers into learning objectives, given the set of training samples $\{(\mathbf x_i, y_i)\}$, one might write the regularization term as 
\begin{equation}\label{eq:framework}
\frac{1}{|\mathcal T|}\sum_{i\in \mathcal T} r(-m_{\mathbf x_i, f}) \quad \mathrm{or}\quad \max_{i\in \mathcal T} r(-m_{\mathbf x_i, f}),
\end{equation}
in which $r(\cdot)$ is a monotonically increasing function. It has been discussed that the specific choice of $\max$ (that focuses on the minimum margin solely) and $\sum$ indicates preference in improving robustness or generalization ability~\cite{Yan2019, Wu2019}. The function $r(\cdot)$ is introduced to optionally put more stress on samples with smaller margins. $\mathcal T$ can be used to enlarge margins for correctly classified training samples only, since the margin for incorrectly classified samples are not well defined.
They form the three key components in developing different margin-based regularizers.

Somewhat surprisingly, Wu et al.~\cite{Wu2019} showed that there might exist an interesting trade-off between minimum margins and average margins. It is uncontroversial, at least for SVMs, that maximizing the minimum margin (i.e., choosing the formulation with $\max$ in Eq.~\eqref{eq:framework}) leads to improved generalization ability, while the average-based term in Eq.~\eqref{eq:framework} favors adversarial robustness~\footnote{It has also been shown that in addition to the average margin which characterizes the first-order statistics of the margin distribution, the variance of the margin is also of importance~\cite{zhang2019optimal}}. In fact, we typically use the average magnitude to evaluate the (adversarial) robustness of DNNs~\cite{Moosavi2016, Moosavi2017, Carlini2017, Guo2018}. However, it is also discovered that using an identity function for $r(\cdot)$ as $-\sum_{i\in \mathcal T} m_{\mathbf x_i, f}/|\mathcal T|$ results in poor prediction accuracy on benign inputs. The regularizer is found to be dominated by some ``extremely robust'' samples so that if all the samples are treated equally in the regularizer, it would be difficult to well trade off the benign-set accuracy and adversarial robustness. A nonlinear ``shrinkage'' function that penalize more on samples with small margins can be introduced to relieve the problem~\cite{Yan2018, Stutz2019confidence}. 

Although using $\min$ or $\max$ instead of the average operation in the regularizer aligns better with the generalization ability, it may lead to slower training convergence since at most one (instance-specific) margin in a whole batch of samples is effectively optimized at each training iteration. 
Yan et al.~\cite{Yan2019} proposed to address this issue by incorporating an ``aggregation'' function that aggregates some samples in the batch rather than using only one.

For $\mathcal T$, some methods chose it to be a set of correctly classified samples~\cite{Yan2018, Yan2019, Ding2019} while the others simply used the whole training set. 
For datasets where nearly $100\%$ training accuracy can be obtained, the two options actually lead to similar performance.
 See Figure~\ref{fig:regularizations} for a comparison of different settings in the functions.

\subsection{DNN Certification}\label{sec:certif}
A related and surging category of methods for DNN \textbf{certification} (also known as DNN verification) was also widely explored in the community~\cite{Tjeng2017, Cheng2017maximum, Katz2017reluplex, Dvijotham2018, Wong2018, Singh2018fast, Boopathy2019, Cohen2019, Zhang2018efficient, fromherz2020fast}. 
These DNN certification methods aimed at maximizing the volume of the hyper-sphere centered at each training instance, in which all data points are predicted into the same class. 
In fact, the radius of such a certified hyper-sphere is actually a lower bound of the classification margin, therefore the techniques could also be regarded as opting to encouraging large classification margins. 

The certifications of DNNs are normally rigorous such that the margins and DNN robustness can be theoretically guaranteed.
One of their demerits might be the high computational complexity.
In fact, it is challenging for most of them to be generalized to large networks on large datasets (e.g., ImageNet~\cite{Russakovsky2015}).
For fast certification, Lee et al.~\cite{Lee2019} and Croce et al.~\cite{Croce2019} proposed to expand linear regions where training samples reside. 
The linear regions can be smaller than the certified hyper-spheres, and they are related to the classification margins similarly.
The relationship between margins and the Lipschitz constant is also exploited, to achieve better certification efficiency~\cite{tsuzuku2018lipschitz}.

\section{Data Augmentation and DNN Compression May Affect Margins}

In addition to the methods introduced in Section~\ref{sec:lmdnn}, there exist other methods that possibly achieve large classification margins as some sort of a byproduct. In general, methods that benefit the generalization ability and test accuracy of DNNs may unintentionally enlarge margins to some extent.
From this point of view, it has been discussed under what condition(s) can data augmentation lead to margin improvement~\cite{Rajput2019}. An achieved result is that, for linear classifiers or linearly separable data, polynomially many more samples are required for a very specific data augmentation strategy to obtain optimal margins. 
Adversarial training~\cite{Madry2018} is often also regarded as a data augmentation strategy, and we know that it has to enlarge margin to guarantee performance.
In fact, it has been shown that adversarial training converges to maximum margin solutions faster than normal training~\cite{charles2019convergence}.
Other augmentations include cutout~\cite{Devries2017improved} and mixup~\cite{zhang2017mixup} may also be related with enlarged margins, considering their success in improving the generalization ability of DNNs, and we encourage future work to explore along this direction.

Though it lacks obvious evidence that can demonstrate the relationship between DNN margins and other learning technologies, we would like to discuss directions that can possibly be explored. The first set of technologies that attract our attention is DNN compression, including network pruning~\cite{Han2015learning, Guo2016dynamic, Wen2016learning} and quantization~\cite{Courbariaux2016binarized, Zhou2017incremental}, since improved or at least similar test-set accuracy can be achieved with significantly fewer learnable parameters (in bit for quantization) using these methods~\cite{Han2017dsd, Zhou2017incremental}. It is also discovered that such network compression leads to improved adversarial robustness in certain circumstances~\cite{Guo2018, Galloway2017attacking}. As have been introduced, both the generalization ability and robustness are closely related to the classification margins, thus we conjecture that it is possible that compression along with re-training also help to achieve models with enlarged margins.

\section{Estimation of The Margin}

Classification margins have been used for a variety of goals, \eg, improving and estimating DNN model robustness and generalization ability~\cite{Elsayed2018, Yan2018, Yan2019, jiang2018predicting, Ding2019}. However, it is still an open problem to find an accurate approximation for the DNN classification margin. As mentioned, the radius of a certified hyper-sphere bounds the margin from blow, while the adversarial examples act as upper bounds of the margin. Hence, taking advantage of DNN certification~\cite{Tjeng2017, Cheng2017maximum, Katz2017reluplex, Dvijotham2018, Wong2018, Singh2018fast, Boopathy2019, Cohen2019, Zhang2018efficient, fromherz2020fast} and adversarial attacks~\cite{Carlini2017, Madry2018}, one can reasonably estimate the range where the classification margin resides in. Other methods for estimating the margin, probably not rigorously, can also be found in Section~\ref{sec:lmdnn}.

\section{Conclusion }\label{sec:conclusion}
In this paper, we have surveyed recent research efforts on classification margin for (nonlinear) DNNs. Unlike for SVMs, the studies are more challenging for DNNs on account of their hierarchical structure and complex nonlinearity. We have revisited some work in the last century and highlight the focus of this paper in the first section, and we have then summarized connections between the margin, generalization, and robustness, mostly from a theoretical point of view, which highlights the importance of large margin even in the state-of-the-art DNN models. We have reviewed methods that target at large margin DNNs over the past few years, and we have categorized them into groups, in a comprehensive but summarized manner. We managed to shorten the paper such that crucial spirit of large margin learning and related methods could be better emphasized. We have shared our view on the key components of current winning methods and point to directions that can possibly be explored.

\section*{Acknowledgments}
This work is funded by the National Key Research and Development Program of China (No. 2018AAA0100701) and a grant from the Guoqiang Institute, Tsinghua University.

\bibliographystyle{IEEEtran}
\bibliography{ref}

\begin{IEEEbiography}[{\includegraphics[width=1in,height=1.25in,clip,keepaspectratio]{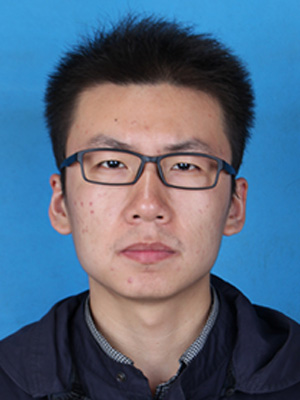}}]{Yiwen Guo} received the B.E. degree from Wuhan University, Wuhan, China, in 2011, and the Ph.D. degree from Tsinghua University, Beijing, China in 2016. He is currently a research scientist at ByteDance AI Lab. Prior to this, he was a staff research scientist at Intel Labs. His current research interests include computer vision, machine learning, and security.
\end{IEEEbiography}

\begin{IEEEbiography}[{\includegraphics[width=1in,height=1.25in,clip,keepaspectratio]{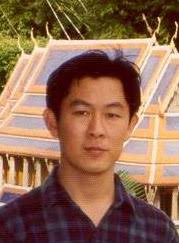}}]{Changshui Zhang} received the B.E. degree in mathematics from Peking University, Beijing, China, in 1986, and the M.S. and Ph.D. degrees in control science and engineering from Tsinghua University, Beijing, in 1989 and 1992, respectively. In 1992, he joined the Department of Automation, Tsinghua University, where he is currently a professor. His research interests include pattern recognition and machine learning. He is a Fellow member of the IEEE.
\end{IEEEbiography}
\end{document}